\documentclass[review]{elsarticle}
\graphicspath{ {./figures/} }
\usepackage{hyperref}
\usepackage{float}
\usepackage{verbatim} 
\usepackage{apalike}
\usepackage{subfig}

\restylefloat{figure}
\restylefloat{table}
\usepackage[T1]{fontenc}
\usepackage{upquote}
\usepackage{mfirstuc}

\journal{Expert Systems with Applications}

\biboptions{authoryear}
\DeclareUnicodeCharacter{2212}{-}
\begin{document}
\begin{frontmatter}

\begin{titlepage}
\begin{center}
\vspace*{1cm}

\textbf{ \large Multimodal Fake News Detection}

\vspace{1.5cm}

Santiago Alonso-Bartolome (santiagoalonsobartolome@gmail.com), Isabel Segura-Bedmar (isegura@inf.uc3m.es)\\

\hspace{10pt}

\begin{flushleft}
\small  
Computer Science Department, Universidad Carlos III de Madrid, Avenida de la Universidad, 30, Leganés, 28911, Madrid, Spain \\

\vspace{1cm}
\textbf{Corresponding Author:} \\
Isabel Segura-Bedmar \\
Computer Science Department, Universidad Carlos III de Madrid, Avenida de la Universidad, 30, Leganés, 28911, Madrid, Spain \\
Tel: +34 916245961 \\
Email: isegura@inf.uc3m.es

\end{flushleft}        
\end{center}
\end{titlepage}

\title{Multimodal Fake News Detection}

\author{Santiago Alonso-Bartolome}
\ead{santiagoalonsobartolome@gmail.com}

\author{Isabel Segura-Bedmar \corref{cor1}}
\ead{isegura@inf.uc3m.es}

\cortext[cor1]{Corresponding author.}
\address{Computer Science Department, Universidad Carlos III de Madrid, Avenida de la Universidad, 30, Leganés, 28911, Madrid, Spain}

\begin{abstract}
Over the last years, there has been an unprecedented proliferation of fake news. As a consequence, we are more susceptible to the pernicious impact that misinformation and disinformation spreading can have in different segments of our society. Thus, the development of tools for automatic detection of fake news plays and important role in the prevention of its  negative  effects. Most attempts to detect and classify false content focus only on using textual information. Multimodal approaches are less frequent and they typically classify news either as true or fake. In this work, we perform  a fine-grained classification of fake news on the Fakeddit dataset, using both unimodal and multimodal approaches. 
Our experiments show that the multimodal approach based on a Convolutional Neural Network (CNN) architecture combining text and image data achieves the best results, with an accuracy of 87\%. Some fake news categories such as Manipulated content, Satire or False connection strongly benefit from the use of images. Using images also improves the results of the other categories, but with less impact.
Regarding the unimodal approaches using only text, Bidirectional Encoder Representations from Transformers (BERT) is the best model with an accuracy of 78\%. 
Therefore, exploiting both text and image data significantly improves the performance of fake news detection. 

\end{abstract}


\end{frontmatter}

\section{Introduction}
\label{introduction}

Digital medial has provided a lot of benefits to our modern society such as 
facilitating social interactions, boosting productivity, improving sharing information, among many others. However, it has also also led to the proliferation of fake news \citep{Finneman_art}, that is, news articles containing false information that has been deliberately created \citep{Allcot_art}. The effects of this kind of misinformation and disinformation spreading can be seen in different segments of our society. The \textit{Pizzagate} incident \citep{Pizzagate} as well as the mob lynchings that occurred in India \citep{India_lynching} are some of the most tragic examples of the consequences of fake news dissemination. Changes in health behaviour intentions \citep{Health_behaviour_effects}, an increase in vaccine hesitancy \citep{Vaccine_hesitancy}, and significant economic losses \citep{economic_loss} are also some of the negative effects that the spread of fake news may have.

Every day, a huge quantity of digital information is produced, making impossible the detection of fake news by manual fact checking. Due to this, it becomes essential to count with techniques that help us to automate the identification of fake news so that more immediate actions can be taken. 
During the last years, several studies have already been carried out to perform an automatic detection of fake news \citep{thota_fake_news, choudhary_fake_news, singh_fake_news, giachanou_fake_news, singhal_fake_news, wang_fake_news}. Most works have focused only on using textual information (unimodal approaches). Much less effort has been devoted to explore multimodal approaches \citep{singh_fake_news, giachanou_fake_news, kumari_fake_news}, which exploit both texts and images to detect the fake news, obtaining better results than the unimodal approaches. 
However, these studies typically address the problem of fake news detection as a binary classification task (that is, consisting on classifying news as either true or fake).

Therefore, the main goal of this paper is to study both unimodal and multimodal approaches to deal with a  finer-grained classification of fake news. To do this, we use the Fakeddit dataset \citep{fakkedit}, made up of posts from Reddit. The posts were classified into the following six different classes: true, misleading content, manipulated content, false connection, imposter content and satire. We explore several deep learning architectures for text classification such as Convolutional neural network (CNN) \citep{convolutional_network}, Bidirectional long short-term memory (BiLSTM) \citep{LSTM_network} and Bidirectional encoder representations from transformers (BERT) \citep{bert_paper}. As multimodal approach, we  propose a CNN architecture that combines both texts and images to classify the fake news. 


\section{Related work}
\label{related_work}

Since the revival of neural networks in the second decade of the current century, many different applications of deep learning techniques have emerged. Part of Natural Language Processing (NLP) and Computer Vision advances are due to the incorporation of deep neural network approaches \citep{related_work_Mahony, related_work_Deng}. Fields such as object recognition \citep{related_work_Zhao}, image captioning \citep{related_work_Hossain}, sentiment analysis \citep{related_work_Tang} or question answering \citep{related_work_Sharma}, among others, have benefited from the development of deep learning in recent years.

Text classification problems are also one of the tasks for which deep neural networks are being extensively used \citep{minaee2021deep}. Most of these works have been based on unimodal approaches that only exploit texts. 
 More ambitious architectures that combine several modalities of data (such as text and image) have also been tried \citep{related_work_Abavisani, related_work_Bae, relatead_work_Yu, related_work_Viana, related_work_Gaspar}. 
 The main intuition behind these multimodal approaches is that  many texts are often accompanied by images, and these images may provide useful information to improve the results of the classification task \citep{related_work_Baheti}. 
 


We now focus on the recent research on fake news detection, distinguishing between unimodal and multimodal approaches. 

\subsection{Unimodal approaches for fake news detection}

We review the most recent studies for the detection of fake news using only the textual content of the news.


\cite{related_work_Wani} use the Constraint@AAAI Covid-19 fake news dataset \citep{related_work_Patwa}, which contains tweets classified as true or fake. 
Several methods are evaluated: CNN, LSTM, Bi-LSTM + Attention, Hierarchical Attention Network (HAN) \citep{related_work_Yang}, BERT, and DistilBERT \citep{related_work_Sanh}, a smaller version of BERT. The best accuracy obtained is 98.41 \% by the DistilBERT model when it is pre-trained on a corpus of Covid-19 tweets. 

\cite{related_work_Goldani} use a capsule network model \citep{related_work_Sabour} based on CNN and pre-trained word embeddings for fake news classification over the ISOT \citep{related_work_Ahmed_ISOT} and LIAR \citep{related_work_Wang_LIAR} datasets. The ISOT dataset is made up of fake and true news articles collected from \textit{Reuters} and \textit{Kaggle}, while  the LIAR dataset contains short statements classified into the following six classes: pants-fire, false, barely-true, half-true, mostly-true and true.
Thus, the authors perform both binary and multi-class fake news classification. The best accuracy obtained with the proposed model is 99.8 \% for the ISOT dataset (binary classification) and 39.5 \% for the LIAR dataset (multi-class classification).

\cite{related_work_Girgis} perform fake news classification using the above mentioned LIAR dataset. More concretely, they use three different models: vanilla Recurrent Neural Network \citep{related_work_Aggrawal_RNN}, Gated Recurrent Unit (GRU) \citep{related_work_Chung} and LSTM. The GRU model obtains an accuracy of 21.7 \%, slightly outperforming the LSTM (21.66 \%) and the vanilla RNN (21.5 \%) models. 

From this review on approaches using only texts, we can conclude that deep learning architectures provide very high accuracy for the binary classification of fake news, however, the performance is much lower when these methods address a fine-grained classification of fake news. Curiously enough, although BERT is reaching state-of-the-art results in many text classification tasks, it has hardly ever used for the multiclassification of fake news.

\subsection{Multimodal approaches for fake news detection}

\cite{singh_fake_news} study the improvement in performance on the binary classification of fake news when textual
and visual features are combined as opposed to using only text or image. They explore several traditional machine learning methods: logistic regression (LR) \citep{logistic_regression}, classification and regression tree (CART) \citep{related_work_Hastie}, linear discriminant analysis (LDA) \citep{related_work_Murphy}, quadratic discriminant analysis (QDA) \citep{related_work_Murphy}, k-nearest neighbors (KNN) \citep{related_work_Murphy} , naïve Bayes (NB) \citep{naive_bayes}, support vector machine (SVM) \citep{suppor_vector_machines} and random forest (RF) \citep{random_forest}. The authors use a Kagle dataset of fake news \citep{related_work_Kaggle_dataset}.
Random forest is the best model with an accuracy of 95.18 \%.

\cite{giachanou_fake_news} propose a model to perform multimodal classification of news articles as either true or fake. In order to obtain textual representations, the BERT model \citep{bert_paper} is applied. For the visual features, the authors use the VGG \citep{related_work_Simonyan} network with 16 layers followed by a LSTM layer and a mean pooling layer. The dataset used by the authors is retrieved from the FakeNewsNet collection \citep{related_work_Shu}. More concretely the authors use 2,745 fake news and 2,714 real news collected from the \textit{GossipCop} posts of such collection. The proposed model achieves an F1-score of 79.55 \%.

Finally, another recent architecture proposed for multimodal fake news classification can be found in the work carried out by \cite{kumari_fake_news}. The authors propose a model that is made up of four modules: i) ABS-BiLSTM (attention based stacked BiLSTM) for extracting the textual features, ii) ABM-CNN-RNN (attention based CNN-RNN) to obtain the visual representations, ii) MFB (multimodal factorized bilinear pooling), where the feature representations obtained from the previous two modules are fused, and iv) MLP (multi-layer perceptron), which takes as input the fused feature representations provided by the MFB module, and then generates the probabilities for each class (true of fake). In order to evaluate the model, two datasets are used: Twitter \citep{related_work_Boididou} and Weibo \citep{related_work_Jin}. The Twitter dataset contains tweets along with images and contextual information. 
The Weibo dataset is made up of tweets, images and social context information. 
The model obtains an accuracy of 88.3\% on the Twitter dataset and an accuracy of 83.2\% on the Weibo dataset .

Apart from the previous studies, several authors have proposed fake news classification models and have evaluated them using the Fakeddit dataset. \cite{related_work_Kaliyar} propose the DeepNet model for a binary classification of fake news. This model is made up of one embedding layer, three convolutional layers, one LSTM layer, seven dense layers, ReLU for activation and finally the softmax function for the binary classification. The model is evaluated on the Fakeddit and BuzzFeed \citep{related_work_Kaggle_BuzzFeed} datasets. The BuzzFeed dataset contains news articles collected within a week before the U.S. election and they are classified as either true or fake. The models provides an accuracy of 86.4\% on the Fakeddit dataset (binary classification) and 95.2\% over the BuzzFeed dataset.

\cite{related_work_Kirchknopf} use four different modalities of data to perform binary classification of fake news over the Fakeddit dataset. More concretely, the authors use the textual content of the news, the associated comments, the images and the remaining metadata belonging to other modalities. The best accuracy obtained is 95.5\%. \cite{related_work_Li2} propose the Entity-Oriented Multi-Modal Alignment and Fusion Network (EMAF) for binary fake news detection. The model is made up of an encapsulating module, a cross-modal alignment module, a cross-model fusion module and a classifier. The authors evaluate the model on the Fakeddit, Weibo and Twitter datasets obtaining accuracies of 92.3\%, 97.4\% and 80.5 \%, respectively. 

\cite{related_work_Xie} propose the Stance Extraction and Reasoning Network (SERN) to obtain stance representations of a post and its associated reply. They combine these stance representations with a multimodal representation of the text and image of a post in order to perform binary fake news classification. The authors use the PHEME dataset \citep{related_work_Zubiaga_PHEME} and a reduced version of the Fakeddit dataset created by them. The PHEME dataset contains 5,802 tweets of which 3,830 are real and  1,972 are false. The accuracies obtained are 96.63 \% (Fakeddit) and 76.53 \% (PHEME).

Finally, \cite{related_work_Kang} use an heterogeneous graph named News Detection Graph (NDG) that contains domain nodes, news nodes, source nodes and review nodes. Moreover, they propose an Heterogeneous Deep Convolutional Network (HDGCN) in order to obtain the embeddings of the news nodes in NDG. The authors evaluate this model using reduced versions of the Weibo and Fakeddit datasets. For the Weibo dataset, they obtain an F1 score of 96 \%, while for the Fakeddit dataset they obtain F1 scores of 88.5\% (binary classification), 85.8 \% (three classes) and 83.2\% (six classes).

As we can see from this review, most multimodal approaches that were evaluated on the Fakkeddit dataset only address the binary classification of fake news. Only one of them \citep{related_work_Kang} has dealt with the multi-classification of fake news using a reduced version of this dataset.

To the best of our knowledge, this is the first work that addresses a fine-grained classification of fake news using the whole Fakeddit dataset. 
Furthermore, contrary to the work proposed in \citep{related_work_Kang}, which exploits a deep convolutional network, we propose a multimodal approach that simply uses a CNN, obtaining very similar performance.


\section{Methods}
\label{methods}


\subsection{Unimodal approaches}

Three unimodal models only using the texts are proposed: CNN, BiLSTM and BERT. 

\subsubsection{Preprocessing for deep learning models}
\label{deep_learning_preprocessing}

We start pre-processing the documents in the corpus by removing stopwords, punctuations, numbers and multiple spaces. Then, we split each text into tokens and we apply lemmatization. After lemmatization, we transform the texts into sequences of integers. This is done firstly by learning the vocabulary of the corpus and building a dictionary where each word is mapped to a different integer number. Then, this dictionary is used to transform each text into a sequence of integers. Every non-zero entry in such sequence corresponds to a word in the original text. The 
original order of the words in the text is respected.

\begin{figure}[H]
    \centering
    \includegraphics[scale = 0.5]{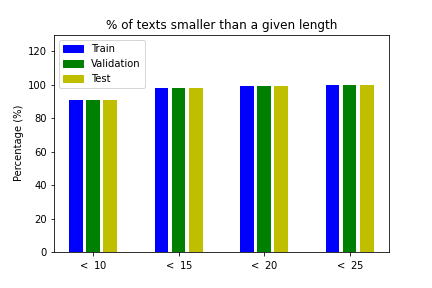}
    \caption{\% of texts smaller than a given length}
    \label{fig:figure_1}
\end{figure}

As we need to feed the deep learning models with vectors of the same length, we pad and truncate the sequences of integers so that they have the same number of entries.
This has the disadvantage that those vectors that are too long will be truncated and some information will be lost. 
In order to select the length of the padded\textbackslash truncated vectors, we computed the percentage of texts that are shorter than 10, 15, 20 and 25 tokens.
Figure \ref{fig:figure_1} shows the results for the training, validation and test partitions in each case. We can see that 98 \% of the texts are smaller than 15 in length. Since the number of texts that will have to be truncated is very small (less than 2 \%) very little information is lost. Therefore, we selected 15 as the length of the vectors after padding and truncating. 

The deep learning architectures use the sequence of word embedddings corresponding to a given text as input. For this reason, an embedding layer will transform each integer value from the input sequence into a vector of word embeddings. In this way, every vectorized document is transformed into matrix of 15 rows and 300 columns (300 being the dimension of the word embeddings). We use both random initialization and pre-trained Glove word embeddings \citep{methods_GloVe}. We also compare a dynamic approach (letting the model further train the word embeddings) and a static approach (not
letting the model train the word embeddings). 

\subsubsection{CNN}
\label{CNN_unimodal}

We now explain the CNN architecture for the text classification of fake news. As was mentioned above, the first layer is an embedding layer. We initialize the embedding matrix using both random initialization and the pre-trained GloVe word embeddings of dimension 300. We chose this size for the word embeddings over other options (50, 100 or 200) because word embeddings of a larger dimension have been proven to give better results \citep{methods_Patel}.

After the embedding layer, we apply four different filters in a convolutional layer. Each of these filters slides across the (15 x 300) matrix with the embeddings of the input sequence and generates 50 output channels. The 4 filters have sizes (2 x 300), (3 x 300), (4 x 300) and (5 x 300), respectively, since these are the typical filter sizes of a CNN for text classification \citep{methods_Voita}. As a consequence the outputs of the previous filters have shapes (14 x 1), (13 x 1), (12 x 1) and (11 x 1), respectively.

The next step is to pass the outputs obtained from the previous layer through the ReLU activation function. This function is applied elemen-twise, and therefore, it does not alter the size of the outputs obtained after the previous step. The effect of this function is to set all the negative values to 0 and leave the positive values unchanged.

After going through the ReLU activation, a maxpooling layer is applied that selects the biggest element out of each of the 200 feature maps (50 feature maps per each of the 4 filters). Thus, 200 single numbers are generated.

These 200 numbers are concatenated and the result is passed through 2 dense layers with one ReLU activation in between \citep{methods_Minaee}. The resulting output is a vector of six entries (each entry corresponding to a different class of the Fakeddit dataset) that, after passing through the logsoftmax function, can be used to obtain the predicted class for the corresponding input text.

\begin{figure}[H]
    \centering
    \includegraphics[scale = 0.65]{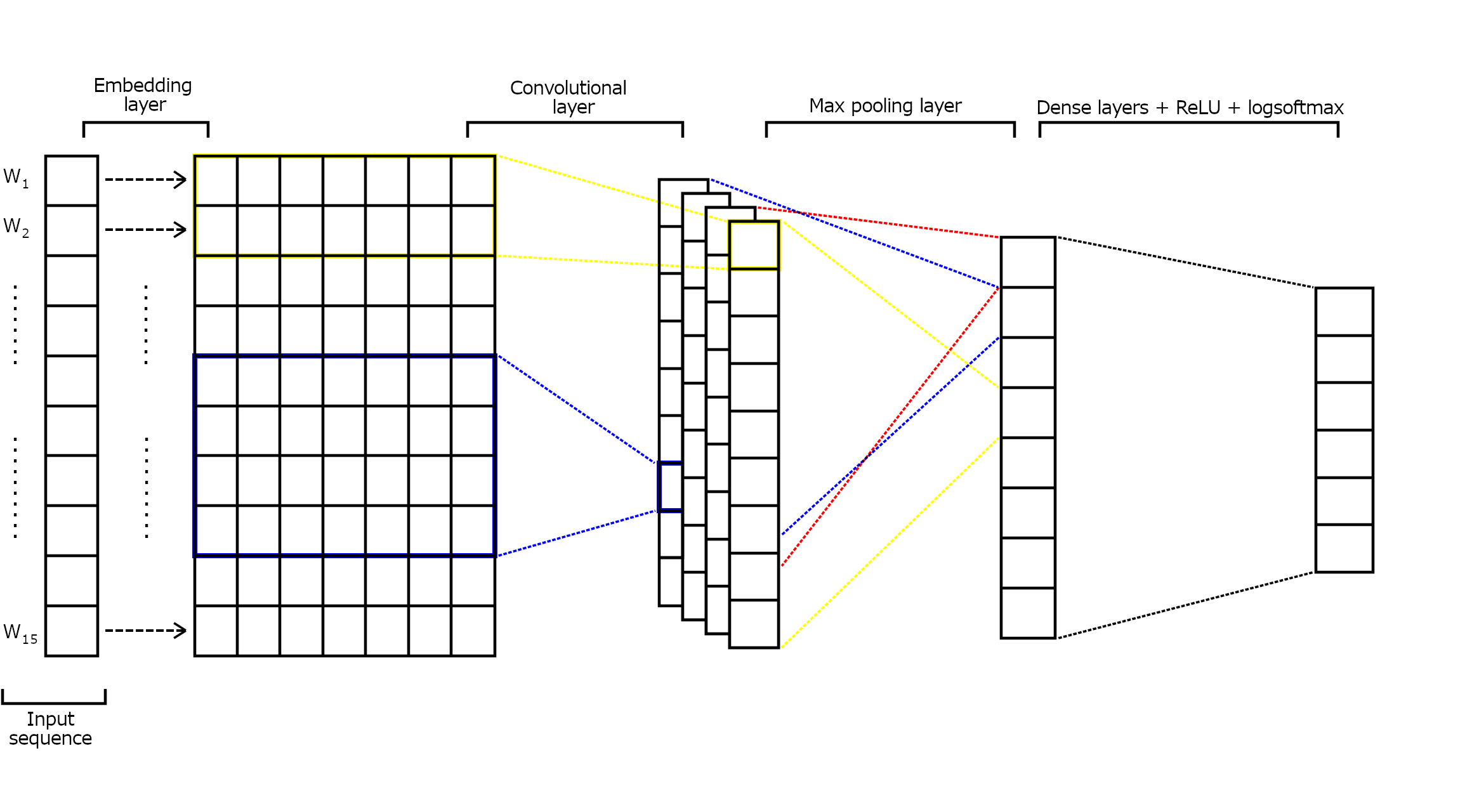}
    \caption{CNN for text classification.}
    \label{fig:figure_2}
\end{figure}

Early stopping \citep{methods_Brownlee} with the train and validation partitions is used in order to select the appropiate number of epochs. We use the \textit{Adam} optimization algorithm \citep{methods_Kingma} for training the model and the \textit{negative log likelihood} as loss function. Figure \ref{fig:figure_2} shows the CNN architecture for text classification.

\subsubsection{BiLSTM}

Actually, this model is a hybrid model that uses a bidirectional LSTM followed by a CNN layer.
Firstly, texts are processed as was described above and these input are passed through the same embedding layer that used for the CNN model. Therefore, each input vector of length 15 is transformed into a matrix of shape 15 x 300.

Then, the matrix with the word embeddings goes through a bidirectional LSTM layer with hidden states of length 70. The output of this layer is a matrix of size 15 x 140 that contains two hidden states (corresponding to the two directions of the BiLSTM) for each word embedding.
The output of the BiLSTM layer is the input of a convolutional layer, which applies 240 filters of size (3 x 140) therefore it generates 240 output arrays of size (13 x 1).
Then, the ReLU activation is applied followed by a maxpooling layer that selects the largest element within each of the 240 feature maps. Thus, this layer outputs a sequence of 240 numbers.

\begin{figure}[H]
    \centering
    \includegraphics[scale = 0.63]{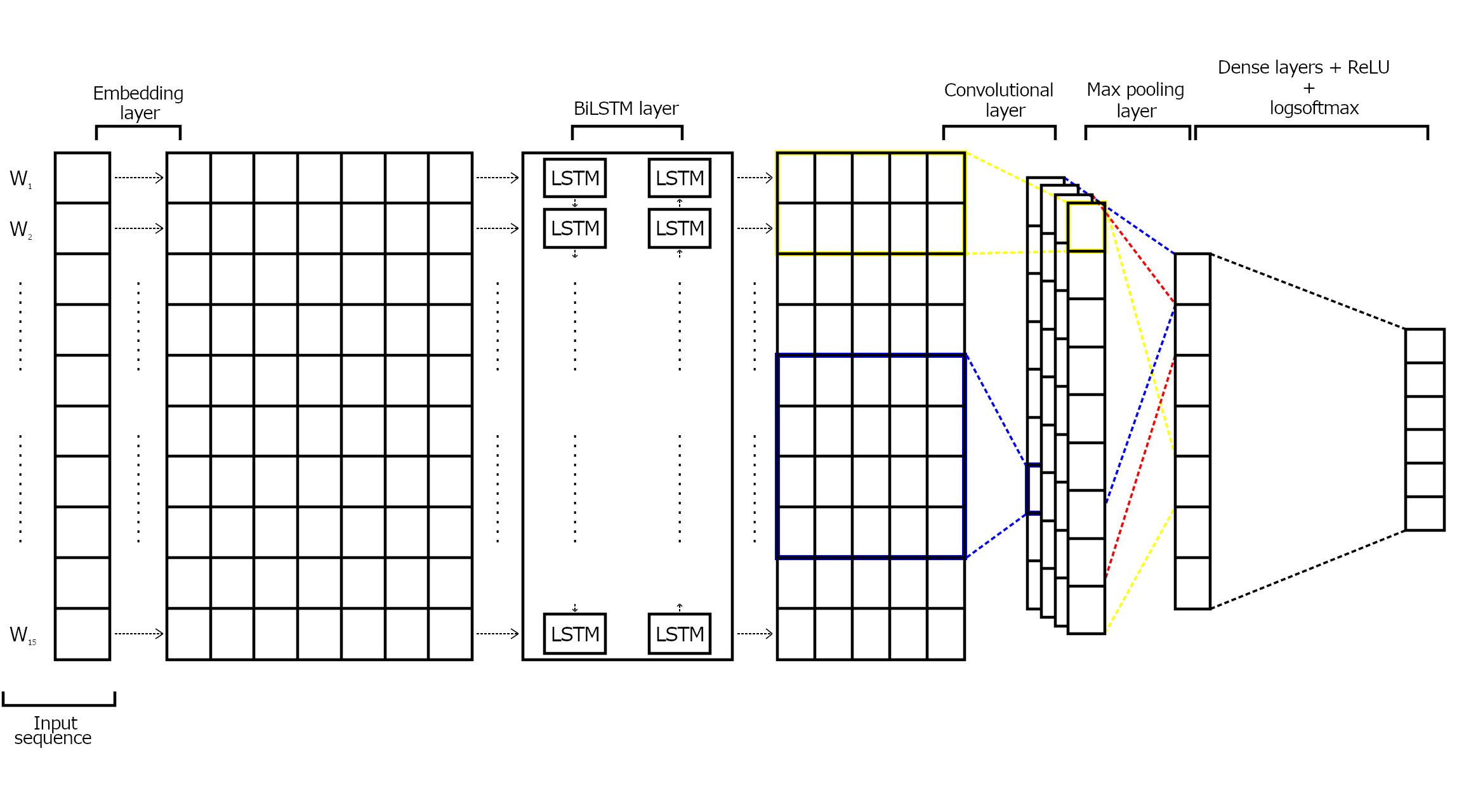}
    \caption{BiLSTM for text classification.}
    \label{fig:figure_3}
\end{figure}

Similarly to what was done for the CNN model, the output of the maxpooling layer is concatenated and passed through two dense layers with  ReLU activation in betweeen. The resulting vector goes through the logsoftmax function and the predicted class in obtained.
Figure \ref{fig:figure_3} shows the architecture of the BiLSTM for text classification.

Early stopping is again used for selecting the optimal number of epochs. We use \textit{Adam} as the optimization algorithm and the \textit{negative log likelihood} as the loss function.

\subsubsection{BERT}

In this case, instead of using random initialization of the pre-traiend Glove embeddings, we now use the vectors provided by BERT to represent the input tokens. As opposed to the GloVe model \citep{methods_GloVe}, BERT is taking into account the context of each word (that is, the words that surround it).

For the pre-processing of the texts the steps are similar to those described above. The main differences are that we tokenize the texts by using the BertTokenizer class from the transformers library \citep{methods_Transformers}. This class has its own vocabulary with the mappings between words and ID’s so it was not necessary to train a tokenizer with the corpus of texts. We also add the [CLS] and [SEP] tokens at the beginning and at the end of each tokenized sequence. It was also necessary to create an attention mask in order to distinguish what entries in each sequence correspond to real words in the input text and what entries are just 0’s resulting from padding the sequences. Thus, the attention mask is composed of 1’s (indicating non-padding entries) and 0’s (indicating padding entries). We use the BERT base model in its uncased version (12 layers, 768
hidden size, 12 heads and 110 million parameters). 

Then, we fine-tune it on our particular problem, that is, the multi-classification of fake news. To do this, we add a linear layer on top of the output of BERT that receives a vector of length 768 and outputs a vector of length 6.

For the training process, we used the \textit{Adam} algorithm for optimization with a learning rate of $2 \cdot 10^{−5}$. We trained the model for two epochs, since the authors of BERT recommended using between two and four epochs for fine-tuning on a specific NLP task \citep{bert_paper}.

\subsection{Multimodal approach}
Our multimodal approach uses a CNN that takes as inputs both the text and the image corresponding to the same news. The model outputs a vector of six numbers out of which the predicted class is obtained. In the following lines, we describe the preprocessing steps applied before feeding the data into the network as well as the architecture of the network.

For the input texts, we apply the same preprocessing steps described in \ref{deep_learning_preprocessing}. Regarding the preprocessing of the images, we only reshaped them so that all have the same shape (560 x 560). Once the pre-processed data is fed into the network, different operations are applied to text and image. The CNN architecture that we use for the texts is the same that we described in \ref{CNN_unimodal}, except for the fact that we eliminate the last 2 dense layers with ReLU activation in between. 

We now describe the CNN model to classify the images. The data first goes through a convolutional layer. Since each image is made up three channels, the number of input cahnnels of this layer is also 3. Besides, it has 6 output channels. Filters of size (5 x 5) are used with stride equal to 1 and no padding. The output for each input image is therefore a collection of 6 matrices of shape (556 x 556).The output of the convolutional layer passes through a nonlinear activation function (ReLU) and then maxpooling is applied with a filter of size (2 x 2) and a stride equal to 2. The resulting output is a set of six matrices of shape (278 x 278). The output from the maxpooling layer passes again through another convolutional layer that has 6 input channels and 3 output channels. The filter size, stride length and padding are the same as those used in the previous convolutional layer. Then the ReLU non-linear activation function and the maxpooling layer are applied again over the feature maps resulting from the convolutional layer. Thus, for a given input (image) we obtain a set of 3 feature maps of shape (137 x 137). Finally, these feature maps are flattened into a vector of length 56307.

The outputs from the operations applied to each text and image are concatenated into a single vector. Then, this vector is passed through 2 dense layers with a ReLU non-linear activation in between. Finally, the logsoftmax function is applied and the logarithm of the probabilities is used in order to compute the predicted class of the given input.

\section{Evaluation}
\label{sec:result}

\subsection{Dataset}
\label{sec:dataset}

In our experiments, we train and test our models using the Fakeddit dataset \citep{fakkedit}, which consists of a collection of posts from Reddit users. It includes texts, images, comments and metadata. The texts are the titles of the posts submitted by users while the comment are made by other users as a answer to a specific post. Thus, the dataset contains over 1 million instances. 

One of the main advantages of this dataset is that it can be used to implement systems capable to perform a finer-grained classification of fake news than the usual binary classification, which only distinguishes between true and fake news. In the Fakeddit dataset, each instance has a label which distinguishes five categories of fake news, besides the unique category of true news. We describe briefly each category:

\begin{itemize}
    \item True: this category indicates that the news is true.
       \item Manipulated Content: in this case, the content has been manipulated  by different means (such as photo editing, for example).
       
        \item False Connection: this category corresponds to those samples in which the text and the image are not in accordance.
        
    \item Satire \textbackslash Parody: this category refers to those news in which the meaning of the content is twisted or misinterpreted in a satirical or humorous way.
    
    \item Misleading Content: this category corresponds to those news in which the information has been deliberately manipulated or altered in order to mislead the public.
    
    \item Imposter Content: in the context of this project, all the news that belong to this category include content generated by bots.

\end{itemize}



The Fakeddit dataset is divided into training, validation and test partitions. Moreover,  there are two different versions of the dataset: the unimodal dataset, whose instances only contains texts, and the multimodal dataset, whose instances have both text and image. Actually, all texts of the multimodal dataset are also included in the unimodal dataset. 

\begin{figure}[H]

\centering
\subfloat[]{
  \includegraphics[width=65mm]{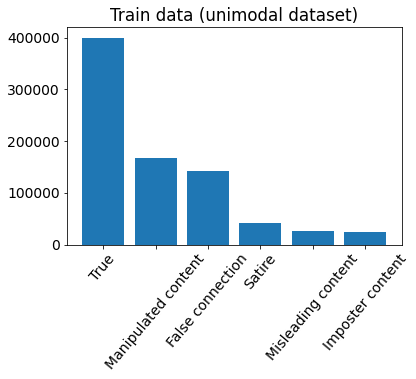}
}
\subfloat[]{
  \includegraphics[width=65mm]{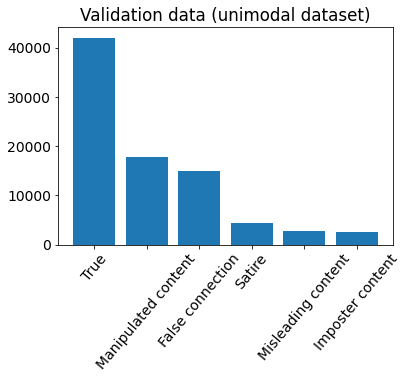}
}
\hspace{0mm}
\subfloat[]{
  \includegraphics[width=65mm]{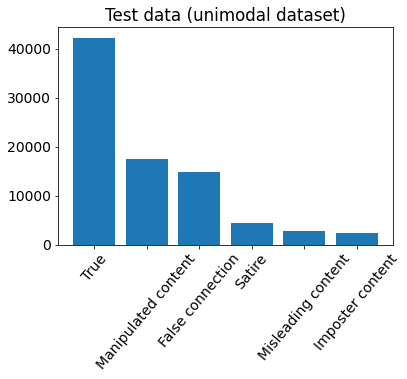}
}

\caption{{\normalfont Class distribution (unimodal dataset). } }
\label{fig:unimodaldataset}
\end{figure}

\begin{figure}[H]
\centering
\subfloat[]{
  \includegraphics[width=65mm]{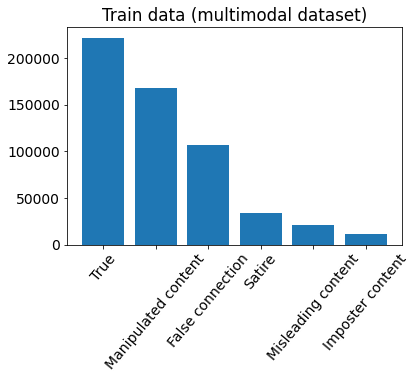}
}
\subfloat[]{
  \includegraphics[width=65mm]{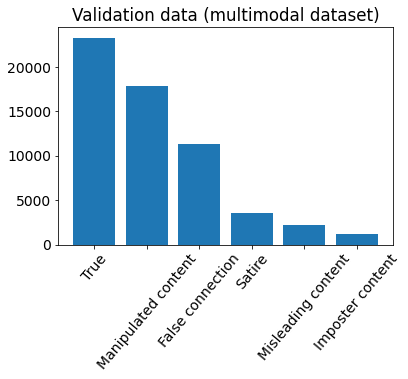}
}
\hspace{0mm}
\subfloat[]{
  \includegraphics[width=65mm]{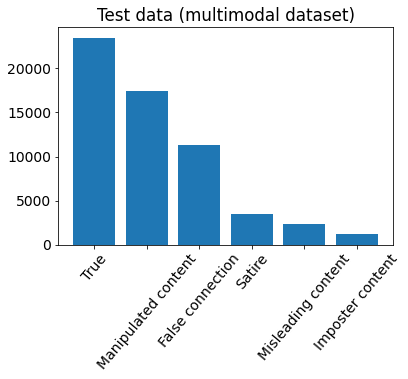}
}

\caption{{\normalfont Class distribution (multimodal dataset). } }
\label{fig:multimodaldataset}
\end{figure}

Figure \ref{fig:unimodaldataset}  shows the distribution of the classes in the unimodal dataset. Figure \ref{fig:multimodaldataset}  provides the same information for the multimodal dataset. 
As we can see, all classes follow a similar distribution in both versions of the dataset (unimodal and multimodal) as well as in the training, validation and test splits. Moreover, both datasets, unimodal and multimodal, are clearly imbalanced (the classes \emph{true}, \emph{manipulated content} and \emph{false connection}  have more instances than the other classes \emph{satire}, \emph{misleading content} and \emph{imposter content}, which are much more underrepresented in both datasets). This imbalance may cause the classification task to be more difficult for those classes with less instances.

\subsection{Results}

In this section, we present the results obtained for each model. We report the \emph{recall}, \emph{precision} and \emph{F1} scores obtained by all the models for each class. The \emph{accuracy} is computed over all the classes. It helps us to compare  models and find the best approach. Moreover, we are also interested in knowing which model is better at detecting only those news  containing false content. For this reason, we also compute the \emph{micro} and \emph{macro} averages of the \emph{recall}, \emph{precision} and \emph{F1} metrics only over five classes of fake news without the  \emph{true} news. 
We use the  $F1_{micro}$ score and the accuracy to compare the performance of the models.


\subsubsection{Unimodal approaches}

\paragraph{\bf CNN results}\mbox{} \\

\begin{table}[H]
\centering
\caption{Results of CNN with random initialization}
\begin{tabular}{lccc}
 \hline
   \textbf{Class} & \textbf{P} & \textbf{R} & \textbf{F1}\\\hline
True & 0.71 & 0.87 & 0.79\\
Manipulated content & 0.75 & 0.84 & 0.79\\
False connection& 0.70 & 0.48 & 0.57\\
Satire & 0.63 & 0.26 & 0.37 \\
Misleading content & 0.71 & 0.54 & 0.61 \\ 
Imposter content& 0.72 & 0.07 & 0.13\\
\hline
  micro-average & 0.73 & 0.62 & 0.57 \\
  macro-average & 0.70 & 0.44 & 0.49 \\\hline
  accuracy & \multicolumn{3}{c}{0.72}\\
  \hline
 \hline
 \end{tabular}
  \label{tab:CNNrandom}
 \end{table}
 
Our first experiment with CNN uses random initialization to initialize the weights of the embedding layer, which are updated during the training process. This model obtains an accuracy of 72\%, a micro F1 of 57\% and a macro F1 of 49\% (see Table \ref{tab:CNNrandom}).
We can also see that True and Manipulated content are the classes with the highest F1 (79\%). A possible reason for this could be that they are the majority classes. On the other hand, the model obtains the lowest F1 (13\%) for Imposter content, which is the minority class in the dataset (see Fig. \ref{fig:unimodaldataset}). Therefore, the results for the different classes appear to be related with the number of instances per class. However, the model achieves an F1 or 61\% for the second minority class, Misleading content. As was explained before, the content of this news has been deliberately manipulated. Identifying these  manipulations appears to be easier than detecting humour or sarcasm in news (Satire) or fake news generated by bots (Imposter content). 

Interestingly, although the model only exploits the textual content of the news, it achieves an F1 of 57\% for classifying the instances of False connection. In these instances, the text and the image are not in accordance.

 

We also explore CNN with static (see Table \ref{tab:CNNstaticGlove}) and dynamic (see Table \ref{tab:CNNdynamicGlove}) GloVe embeddings \cite{methods_GloVe}. In both models, the embedding layer is initialized with the pretrained Glove vectors. When dynamic training is chosen, these vectors are updated during the training process. On the other hand, if static training is chosen, the vectors are fixed during the training process. The model with dynamic vectors  overcomes the one with static vectors, with an slightly improvement in accuracy (roughly one percentage point). However, in terms of micro F1, the static model is better than the dynamic one. Both models provide the same macro F1 (69\%). Regarding the classes, there are not significant differences, except for Imposer content. For this class, updating the pre-trained Glove vectors results in a decrease of seven percentage points in F1.

 \begin{table}[H]
 \centering
\caption{Results of CNN with static Glove vectors}
\begin{tabular}{lccc}
 \hline
   \textbf{Class} & \textbf{P} & \textbf{R} & \textbf{F1}\\\hline
  True & 0.76 & 0.81 & 0.79\\
  Manipulated content & 0.75 & 0.82 & 0.79\\
  False connection& 0.65 & 0.59 & 0.62\\
  Satire & 0.60 & 0.40 & 0.48 \\
  Misleading content & 0.71 & 0.59 & 0.64 \\ 
  Imposter content& 0.35 & 0.21 & 0.26\\
\hline
  micro-average & 0.70 & 0.67 & 0.69 \\
  macro-average & 0.61 & 0.52 & 0.56 \\\hline
  accuracy & \multicolumn{3}{c}{0.73}\\
  \hline
 
 \hline
 \end{tabular}
  \label{tab:CNNstaticGlove}

 \end{table}

\begin{table}[H]
\centering
\caption{Results of CNN with dynamic Glove vectors}
\begin{tabular}{lccc}
 \hline
   \textbf{Class} & \textbf{P} & \textbf{R} & \textbf{F1}\\\hline
  True & 0.74 & 0.87 & 0.80\\
    Manipulated content & 0.76 & 0.83 & 0.80\\
  False connection& 0.71 & 0.54 & 0.61\\

  Satire & 0.67 & 0.35 & 0.46 \\
  Misleading content & 0.74 & 0.58 & 0.65 \\ 
  Imposter content& 0.70 & 0.11 & 0.19\\
\hline
  micro-average & 0.74 & 0.65 & 0.69 \\
  macro-average & 0.71 & 0.48 & 0.54 \\\hline
  accuracy & \multicolumn{3}{c}{0.74}\\
  \hline
 
 \hline
 \end{tabular}
   \label{tab:CNNdynamicGlove}
 \end{table}

We also compare the effect of the pretrained Glove vectors with random initialization (see Table \ref{tab:CNNrandom}). In both dynamic and static approaches, initializing the model with the pretrained GloVe word embeddings gets better results than random initilization. The reason for this is that the GloVe vectors contain information about the relationship between different words that random vectors can not capture.

As the dataset is highly unbalanced, we use the micro F1 to assess and compare the overall performances of the three models. Thus, the best model is a CNN with dynamic Glove vectors. However, the dynamic training takes much more time than the static training (around 6000 to 8000 seconds more). This is due to the fact that, in a dynamic approach,  word embeddings are also learned and this increases significantly the training time.

\paragraph{\bf BiLSTM results}\mbox{} \\
As a second deep learning model, we explore a BiLSTM model. We replicate the same experiments as described for CNN, that is, using random initialization and pre-trained Glove vectors.

\begin{table}[H]
\centering
\caption{Results of BiLSTM with random initialization}
\begin{tabular}{lccc}
 \hline
   \textbf{Class} & \textbf{P} & \textbf{R} & \textbf{F1}\\\hline
  True & 0.70 & 0.88 & 0.78\\
  Manipulated content & 0.73 & 0.85 & 0.79\\
  False connection& 0.73 & 0.44 & 0.55\\
  Satire & 0.58 & 0.25 & 0.35 \\
  Misleading content & 0.71 & 0.54 & 0.61 \\ 
  Imposter content& 0.86 & 0.08 & 0.14\\
  
  \hline
  micro-average & 0.73 & 0.61 & 0.66 \\
  macro-average & 0.74 & 0.41 & 0.48 \\\hline
  accuracy & \multicolumn{3}{c}{0.72}\\
  \hline
 
 \hline
 \end{tabular}
  \label{tab:BiLSTMrandom}
 \end{table}

The BiLSTM initializated with random vectors obtains very similar results to those achieved by CNN with random initialization (see Table \ref{tab:CNNrandom}). In fact, both models provides the same accuracy of 0.72. However, in terms of micro F1, the BiLSM model obtains up to 9 points more than the CNN model with random initialization. This improvement may because the BiLSTM improved its scores for Imposter content.

  

 The use of static Glove vectors appears to have a positive effect on the performance of the BiLSTM model (see Table \ref{tab:BiLSTMstatic}. The model shows significant improvements for False connection, Satire, Misleading content, and Imposter, with  increases of 6, 12, 3, and 10 points, respectively. Therefore, the pretrained Glove vectors get better results than random initialization. 
 
 \begin{table}[H]
 \centering
\caption{Results of BiLSTM with static Glove vectors}
\begin{tabular}{lccc}
 \hline
   \textbf{Class} & \textbf{P} & \textbf{R} & \textbf{F1}\\\hline
  True & 0.74 & 0.85 & 0.79\\
  Manipulated content & 0.77 & 0.82 & 0.79\\
  False connection& 0.68 & 0.55 & 0.61\\

  Satire & 0.55 & 0.41 & 0.47 \\
  Misleading content & 0.77 & 0.55 & 0.64 \\ 
  Imposter content& 0.45 & 0.17 & 0.24\\
\hline
  micro-average & 0.72 & 0.65 & 0.69 \\
  macro-average & 0.65 & 0.50 & 0.55 \\\hline
  accuracy & \multicolumn{3}{c}{0.73}\\
  \hline
 
 \hline
 \end{tabular}
  \label{tab:BiLSTMstatic}

 \end{table}
 
If the pretrained Glove vectors are updated during the training of the BiLSTM model, an accuracy of 75\% is achieved, that is, two points more than BiLSTM with static Glove vectors. Moreover, this models with dynamic Glove vectors improves the results for all classes, with increases ranging from 1 to four points. 
In terms of micro F1, using dynamic Glove vectors is the best approach for BiLSTM. Moreover, this model slightly overcomes the CNN model with dynamic Glove vectors, roughly one percentage point. However, as mentioned above, the dynamic training takes much more time than static training.

 \begin{table}[H]
\centering
\caption{Results of BiLSTM with dynamic  Glove vectors}
\begin{tabular}{lccc}
 \hline
   \textbf{Class} & \textbf{P} & \textbf{R} & \textbf{F1}\\\hline
  True & 0.75 & 0.86 & 0.80\\
Manipulated content & 0.77 & 0.84 & 0.80\\
  False connection& 0.72 & 0.55 & 0.63\\

  Satire & 0.63 & 0.41 & 0.50 \\
  Misleading content & 0.78 & 0.57 & 0.66 \\
  Imposter content& 0.57 & 0.18 & 0.28\\
  
   \hline
  micro-average & 0.74 & 0.67 & 0.70 \\
  macro-average & 0.69 & 0.51 & 0.57 \\\hline
  accuracy & \multicolumn{3}{c}{0.75}\\
  \hline
 
 \hline
 \end{tabular}
   \label{tab:BiLSTMdG}
 \end{table}

\paragraph{\bf BERT results}\mbox{} \\

 \begin{table}[H]
\centering
\caption{BERT results}
\begin{tabular}{lccc}
 \hline
   \textbf{Class} & \textbf{P} & \textbf{R} & \textbf{F1}\\\hline
  True & 0.81 & 0.86 & 0.83\\
    Manipulated content & 0.80 & 0.86 & 0.83\\
  False connection& 0.72 & 0.64 & 0.68\\

  Satire & 0.70 & 0.53 & 0.61 \\
  Misleading content & 0.77 & 0.70 & 0.73 \\ 
    Imposter content& 0.61 & 0.28 & 0.38\\

  \hline
  micro-average & 0.76 & 0.73 & 0.74 \\
  macro-average & 0.72 & 0.60 & 0.65 \\\hline
  accuracy & \multicolumn{3}{c}{0.78}\\
  \hline
 
 \hline
 \end{tabular}
   \label{tab:BERT}
 \end{table}

Table \ref{tab:BERT} shows that BERT achieves an accuracy of  78\% and a micro F1 of 74\%. Therefore, it outperforms all the previous unimodal deep learning approaches.  This proves the advantage of the pre-trained contextual text representations provided by BERT, as opposed to the context-free GloVe vectors or random initialisation for neural networks. 

Moreover, BERT is better in all classes. Comparing the classes, the behaviour of BERT is very similar to the previous deep learning models, that is, the more training instances for a class, the better preditions for it. 
In this way, True and Manipulated content both get the highest F1 (83\%), while the worst performing class is Imposter content (F1=38\%).  
As in previous models, Misleading content gets better scores than Satire, despite the fact that this class is more represented than the first one, Misleading content  (see Fig. \ref{fig:multimodaldataset}). 

\subsubsection{Multimodal approach}

 \begin{table}[H]
\centering
\caption{Multimodal approach results}
\begin{tabular}{lccc}
 \hline
   \textbf{Class} & \textbf{P} & \textbf{R} & \textbf{F1}\\\hline
  True & 0.85 & 0.88 & 0.86\\
  Manipulated content & 1 & 1 & 1\\
  False connection& 0.77 & 0.76 & 0.76\\

  Satire & 0.82 & 0.72 & 0.77 \\
  Misleading content & 0.75 & 0.79 & 0.77 \\ 
  Imposter content& 0.46 & 0.25 & 0.32\\
  
\hline
  micro-average & 0.88 & 0.86 & 0.87 \\
  macro-average & 0.76 & 0.70 & 0.72 \\\hline
  accuracy & \multicolumn{3}{c}{0.87}\\
  \hline
 
 \hline
 \end{tabular}
   \label{tab:multimodal}
 \end{table}

Table \ref{tab:multimodal} shows that the multimodal approach obtains an accuracy of 87\% and a micro F1 of 72\%, which are the highest scores out of all the unimodal models. 

As expected, the training set size for each class strongly affects the model scores. While True and Manipulated content, the majority classes, get the highest scores, Imposter content, the minority class, shows the lowest F1 (32\%), even six points lower than that provided by BERT for the same class (F1=38\%). Thus, we can say that the image content provides little  information for identifying instances of \emph{Imposter content}.
Manipulated content shows an F1 or 100\%. This is probably due to the fact that the images in this category, have  been manipulated. These manipulations may be easily detected by CNN. 

As expected, the use of images significantly  improves the results for False connection. The multimodal model shows an F1 of 76\%, 8 points up than that obtained by BERT, the best unimodal approach, and 15 points up than the unimodal CNN model using only texts. The improvement is even greater for detecting instances of Satire, with an increase of 16 points up than those obtained by BERT and by the unimodal CNN model.

\subsubsection{Comparison of the best models}

 In addition to the deep learning algorithms, we also propose as baseline a Support Vector Machine (SVM), one of the most successful algorithm for text classification. 
Table \ref{tab:bestmodelsACC} shows a comparative of the best models (traditional algorithms, CNN, BiLSTM, BERT and multimodal CNN) according to their accuracy and micro average scores.

\begin{table}[H]
\centering
\begin{tabular}{lcccc}
 \hline
  Model & P & R &  F1 &  Acc.\\
 \hline
 SVM  & 0.71  & 0.64 &  0.67 &  0.72 \\
 CNN (Dynamic + GloVe)   & 0.74   & 0.65 &  0.69 &  0.74 \\
 BiLSTM (Dynamic + GloVe)  & 0.74   & 0.67 &  0.70 & 0.75 \\
 BERT   & 0.76  & 0.73 &  0.74 &  0.78 \\
 Multimodal CNN   & 0.88   & 0.86 & 0.87 &  \textbf{0.87} \\
 \hline
\end{tabular}
 \caption{Comparison of the best models (micro\_averages).}
 \label{tab:bestmodelsACC}
\end{table}

In conclusion, we can see that the multimodal CNN outperforms all the unimodal approaches. This proves the usefulness of combining texts and images for a fine-grained fake news classification. Focusing on the unimodal approaches, the BERT model is the best both in terms of accuracy and micro F1 score, which shows the advantage of using contextual word embeddings. The third best approach is the BiLSTM with dynamic Glove vectors.  Finally, all the deep learning approaches outperform our baseline SVM.

\section{Conclusions}

Fake news could have a significant negative effect on politics, health and  economy. Therefore, it becomes necessary to develop tools  that allow for a rapid and reliable detection of misinformation.

Apart from the work carried out by the creators of the Fakeddit dataset \citep{fakkedit}, this is, to the best of our knowledge, the only study that addresses a fine-grained classification of fake news by performing a comprehensive comparison of unimodal and multimodal approaches based on the most advanced deep learning techniques. 

 The multimodal approach overcomes the approaches that only exploit texts. BERT is the best model for the taks of text classification. Moreover, using dynamic GloVe word embeddings outperforms random initialization for the CNN and BiLSTM architectures.

As future work, we plan to use pre-trained networks to generate the visual representations. 
In particular, we will use the network VGG
, which was  pre-trained on a large dataset of images such as \emph{ImageNet}. We also plan to explore different deep learning techniques such as LSTM, BiLSTM, GRU or BERT, as well as, different methods to combine the visual and textual representations. 
In our current study, we have built our multimodal CNN using an early fusion approach, which consists on creating textual and visual representations, combining then, and then applying a classifier over the resulting combined representation to get the probabilities for each class. Instead of this, we plan to study a late fusion approach, which would require two separate classifiers (one for the textual inputs and the other for the image inputs). The predictions from both classifiers are then combined and the final prediction is obtained.

\section*{Acknowledgements}
This work was supported by the Madrid Government (Comunidad de Madrid) under the Multiannual Agreement with UC3M in the line of "Fostering Young Doctors Research" (NLP4RARE-CM-UC3M) and in the context of the V PRICIT (Regional Programme of Research and Technological Innovation) and under the Multiannual Agreement with UC3M in the line of "Excellence of University Professors (EPUC3M17).


\begin{thebibliography}{67}
\expandafter\ifx\csname natexlab\endcsname\relax\def\natexlab#1{#1}\fi
\providecommand{\url}[1]{\texttt{#1}}
\providecommand{\href}[2]{#2}
\providecommand{\path}[1]{#1}
\providecommand{\DOIprefix}{doi:}
\providecommand{\ArXivprefix}{arXiv:}
\providecommand{\URLprefix}{URL: }
\providecommand{\Pubmedprefix}{pmid:}
\providecommand{\doi}[1]{\href{http://dx.doi.org/#1}{\path{#1}}}
\providecommand{\Pubmed}[1]{\href{pmid:#1}{\path{#1}}}
\providecommand{\bibinfo}[2]{#2}
\ifx\xfnm\relax \def\xfnm[#1]{\unskip,\space#1}\fi
\bibitem[{Abavisani et~al.(2020)Abavisani, Wu, Hu, Tetreault \&
  Jaimes}]{related_work_Abavisani}
\bibinfo{author}{Abavisani, M.}, \bibinfo{author}{Wu, L.}, \bibinfo{author}{Hu,
  S.}, \bibinfo{author}{Tetreault, J.}, \& \bibinfo{author}{Jaimes, A.}
  (\bibinfo{year}{2020}).
\newblock \bibinfo{title}{{Multimodal Categorization of Crisis Events in Social
  Media}}.
\newblock In {\it \bibinfo{booktitle}{2020 IEEE/CVF Conference on Computer
  Vision and Pattern Recognition (CVPR)}\/} (pp.
  \bibinfo{pages}{14667--14677}).
\newblock \bibinfo{address}{Los Alamitos, CA, USA}: \bibinfo{publisher}{IEEE
  Computer Society}.
\bibitem[{Aggarwal(2018)}]{related_work_Aggrawal_RNN}
\bibinfo{author}{Aggarwal, C.~C.} (\bibinfo{year}{2018}).
\newblock \bibinfo{title}{Recurrent neural networks}.
\newblock In {\it \bibinfo{booktitle}{Neural Networks and Deep Learning: A
  Textbook}\/} (pp. \bibinfo{pages}{271--313}).
\newblock \bibinfo{address}{Cham}: \bibinfo{publisher}{Springer International
  Publishing}.
\bibitem[{Ahmed et~al.(2017)Ahmed, Traore \& Saad}]{related_work_Ahmed_ISOT}
\bibinfo{author}{Ahmed, H.}, \bibinfo{author}{Traore, I.}, \&
  \bibinfo{author}{Saad, S.} (\bibinfo{year}{2017}).
\newblock \bibinfo{title}{Detection of online fake news using n-gram analysis
  and machine learning techniques}.
\newblock In {\it \bibinfo{booktitle}{International conference on intelligent,
  secure, and dependable systems in distributed and cloud environments}\/} (pp.
  \bibinfo{pages}{127--138}).
\newblock \bibinfo{organization}{Springer}.
\bibitem[{Bae et~al.(2020)Bae, Park, Lee, Lee \& Lim}]{related_work_Bae}
\bibinfo{author}{Bae, K.~I.}, \bibinfo{author}{Park, J.}, \bibinfo{author}{Lee,
  J.}, \bibinfo{author}{Lee, Y.}, \& \bibinfo{author}{Lim, C.}
  (\bibinfo{year}{2020}).
\newblock \bibinfo{title}{Flower classification with modified multimodal
  convolutional neural networks}.
\newblock {\it \bibinfo{journal}{Expert Systems with Applications}\/},  {\it
  \bibinfo{volume}{159}\/}, \bibinfo{pages}{113455}.
  \DOIprefix\doi{https://doi.org/10.1016/j.eswa.2020.113455}.
\bibitem[{Baheti(2020)}]{related_work_Baheti}
\bibinfo{author}{Baheti, P.} (\bibinfo{year}{2020}).
\newblock \bibinfo{title}{\textit{Introduction to Multimodal Deep Learning}}.
\newblock \bibinfo{note}{Retrieved from
  \url{https://heartbeat.comet.ml/introduction-to-multimodal-deep-learning-630b259f9291}.
  Accessed November 13, 2021}.
\bibitem[{Barber(2012)}]{naive_bayes}
\bibinfo{author}{Barber, D.} (\bibinfo{year}{2012}).
\newblock \bibinfo{title}{Naive bayes}.
\newblock In {\it \bibinfo{booktitle}{Bayesian Reasoning and Machine
  Learning}\/} (pp. \bibinfo{pages}{243--255}).
\newblock \bibinfo{address}{Cambridge}: \bibinfo{publisher}{Cambridge
  University Press}.
\newblock \DOIprefix\doi{10.1017/CBO9780511804779}.
\bibitem[{Boididou et~al.(2015)Boididou, Andreadou, Papadopoulos, Dang-Nguyen,
  Boato, Riegler, Kompatsiaris et~al.}]{related_work_Boididou}
\bibinfo{author}{Boididou, C.}, \bibinfo{author}{Andreadou, K.},
  \bibinfo{author}{Papadopoulos, S.}, \bibinfo{author}{Dang-Nguyen, D.-T.},
  \bibinfo{author}{Boato, G.}, \bibinfo{author}{Riegler, M.},
  \bibinfo{author}{Kompatsiaris, Y.} et~al. (\bibinfo{year}{2015}).
\newblock \bibinfo{title}{{Verifying Multimedia Use at MediaEval 2015}}.
\newblock {\it \bibinfo{journal}{MediaEval}\/},  {\it \bibinfo{volume}{3}\/},
  \bibinfo{pages}{7}.
\bibitem[{Breiman(2001)}]{random_forest}
\bibinfo{author}{Breiman, L.} (\bibinfo{year}{2001}).
\newblock \bibinfo{title}{Random \uppercase{F}orests}.
\newblock {\it \bibinfo{journal}{Machine Learning}\/},  {\it
  \bibinfo{volume}{45}\/}, \bibinfo{pages}{5--32}.
  \DOIprefix\doi{https://doi.org/10.1023/A:1010933404324}.
\bibitem[{Brown(2019)}]{economic_loss}
\bibinfo{author}{Brown, E.} (\bibinfo{year}{2019}).
\newblock \bibinfo{title}{\textit{Online fake news is costing us \$78 billion
  globally each year}}.
\newblock \bibinfo{howpublished}{ZDNet}.
\newblock
  \bibinfo{note}{\url{https://www.zdnet.com/article/online-fake-news-costing-us-78-billion-globally-each-year/}}.
\bibitem[{Brownlee()}]{methods_Brownlee}
\bibinfo{author}{Brownlee, J.} ().
\newblock \bibinfo{title}{{A Gentle Introduction to Early Stopping to Avoid
  Overtraining Neural Networks}}.
\newblock
  \bibinfo{howpublished}{\url{https://machinelearningmastery.com/early-stopping-to-avoid-overtraining-neural-network-models/}}.
\newblock \bibinfo{note}{Accessed October 5, 2021}.
\bibitem[{Choudhary et~al.(2021)Choudhary, Chouhan, Pilli \&
  Vipparthi}]{choudhary_fake_news}
\bibinfo{author}{Choudhary, M.}, \bibinfo{author}{Chouhan, S.~S.},
  \bibinfo{author}{Pilli, E.~S.}, \& \bibinfo{author}{Vipparthi, S.~K.}
  (\bibinfo{year}{2021}).
\newblock \bibinfo{title}{Ber\uppercase{C}onvo\uppercase{N}et: A deep learning
  framework for fake news classification}.
\newblock {\it \bibinfo{journal}{Applied Soft Computing}\/},  {\it
  \bibinfo{volume}{110}\/}, \bibinfo{pages}{107614}.
  \DOIprefix\doi{https://doi.org/10.1016/j.asoc.2021.107614}.
\bibitem[{Chung et~al.(2014)Chung, Gulcehre, Cho \&
  Bengio}]{related_work_Chung}
\bibinfo{author}{Chung, J.}, \bibinfo{author}{Gulcehre, C.},
  \bibinfo{author}{Cho, K.}, \& \bibinfo{author}{Bengio, Y.}
  (\bibinfo{year}{2014}).
\newblock \bibinfo{title}{Empirical evaluation of gated recurrent neural
  networks on sequence modeling}.
\newblock {\it \bibinfo{journal}{arXiv preprint arXiv:1412.3555}\/}, .
\bibitem[{Deng \& Liu(2018)}]{related_work_Deng}
\bibinfo{author}{Deng, L.}, \& \bibinfo{author}{Liu, Y.}
  (\bibinfo{year}{2018}).
\newblock {\it \bibinfo{title}{Deep learning in natural language
  processing}\/}.
\newblock (\bibinfo{edition}{1st} ed.).
\newblock \bibinfo{publisher}{Springer}.
\bibitem[{Devlin et~al.(2018)Devlin, Chang, Lee \& Toutanova}]{bert_paper}
\bibinfo{author}{Devlin, J.}, \bibinfo{author}{Chang, M.-W.},
  \bibinfo{author}{Lee, K.}, \& \bibinfo{author}{Toutanova, K.}
  (\bibinfo{year}{2018}).
\newblock \bibinfo{title}{Bert: Pre-training of deep bidirectional transformers
  for language understanding}.
\newblock {\it \bibinfo{journal}{arXiv preprint arXiv:1810.04805}\/}, .
\bibitem[{Face()}]{methods_Transformers}
\bibinfo{author}{Face, H.} ().
\newblock \bibinfo{title}{{Transformers}}.
\newblock \bibinfo{howpublished}{\url{https://huggingface.co/transformers/}}.
\newblock \bibinfo{note}{Accessed October 5, 2021}.
\bibitem[{Finneman \& Thomas(2018)}]{Finneman_art}
\bibinfo{author}{Finneman, T.}, \& \bibinfo{author}{Thomas, R.~J.}
  (\bibinfo{year}{2018}).
\newblock \bibinfo{title}{A family of falsehoods: Deception, media hoaxes and
  fake news}.
\newblock {\it \bibinfo{journal}{Newspaper Research Journal}\/},  {\it
  \bibinfo{volume}{39}\/}, \bibinfo{pages}{350--361}.
  \DOIprefix\doi{https://doi.org/10.1177/0739532918796228}.
\bibitem[{Gaspar \& Alexandre(2019)}]{related_work_Gaspar}
\bibinfo{author}{Gaspar, A.}, \& \bibinfo{author}{Alexandre, L.~A.}
  (\bibinfo{year}{2019}).
\newblock \bibinfo{title}{A multimodal approach to image sentiment analysis}.
\newblock In {\it \bibinfo{booktitle}{International Conference on Intelligent
  Data Engineering and Automated Learning}\/} (pp. \bibinfo{pages}{302--309}).
\newblock \bibinfo{organization}{Springer, Cham}.
\bibitem[{Giachanou et~al.(2020)Giachanou, Zhang \&
  Rosso}]{giachanou_fake_news}
\bibinfo{author}{Giachanou, A.}, \bibinfo{author}{Zhang, G.}, \&
  \bibinfo{author}{Rosso, P.} (\bibinfo{year}{2020}).
\newblock \bibinfo{title}{Multimodal multi-image fake news detection}.
\newblock In {\it \bibinfo{booktitle}{2020 IEEE 7th International Conference on
  Data Science and Advanced Analytics (DSAA)}\/} (pp.
  \bibinfo{pages}{647--654}).
\newblock \bibinfo{organization}{IEEE}.
\bibitem[{Girgis et~al.(2018)Girgis, Amer \& Gadallah}]{related_work_Girgis}
\bibinfo{author}{Girgis, S.}, \bibinfo{author}{Amer, E.}, \&
  \bibinfo{author}{Gadallah, M.} (\bibinfo{year}{2018}).
\newblock \bibinfo{title}{{Deep Learning Algorithms for Detecting Fake News in
  Online Text}}.
\newblock In {\it \bibinfo{booktitle}{2018 13th International Conference on
  Computer Engineering and Systems (ICCES)}\/} (pp. \bibinfo{pages}{93--97}).
\newblock \DOIprefix\doi{10.1109/ICCES.2018.8639198}.
\bibitem[{Goldani et~al.(2021)Goldani, Momtazi \&
  Safabakhsh}]{related_work_Goldani}
\bibinfo{author}{Goldani, M.~H.}, \bibinfo{author}{Momtazi, S.}, \&
  \bibinfo{author}{Safabakhsh, R.} (\bibinfo{year}{2021}).
\newblock \bibinfo{title}{Detecting fake news with capsule neural networks}.
\newblock {\it \bibinfo{journal}{{Applied Soft Computing}}\/},  {\it
  \bibinfo{volume}{101}\/}, \bibinfo{pages}{106991}.
\bibitem[{Goodfellow et~al.(2016)Goodfellow, Bengio \&
  Courville}]{convolutional_network}
\bibinfo{author}{Goodfellow, I.}, \bibinfo{author}{Bengio, Y.}, \&
  \bibinfo{author}{Courville, A.} (\bibinfo{year}{2016}).
\newblock \bibinfo{title}{Convolutional networks}.
\newblock In {\it \bibinfo{booktitle}{Deep learning}\/} (pp.
  \bibinfo{pages}{321--362}).
\newblock \bibinfo{address}{Cambridge, MA, USA}: \bibinfo{publisher}{MIT
  press}.
\bibitem[{Greene \& Murphy(2021)}]{Health_behaviour_effects}
\bibinfo{author}{Greene, C.~M.}, \& \bibinfo{author}{Murphy, G.}
  (\bibinfo{year}{2021}).
\newblock \bibinfo{title}{Quantifying the effects of fake news on behavior:
  Evidence from a study of \uppercase{COVID}-19 misinformation}.
\newblock {\it \bibinfo{journal}{Journal of Experimental Psychology:
  Applied}\/}, . \DOIprefix\doi{http://dx.doi.org/10.1037/xap0000371}.
\bibitem[{Hastie et~al.(2009{\natexlab{a}})Hastie, Tibshirani \&
  Friedman}]{related_work_Hastie}
\bibinfo{author}{Hastie, T.}, \bibinfo{author}{Tibshirani, R.}, \&
  \bibinfo{author}{Friedman, J.} (\bibinfo{year}{2009}{\natexlab{a}}).
\newblock \bibinfo{title}{Additive models, trees, and related methods}.
\newblock In {\it \bibinfo{booktitle}{The Elements of Statistical Learning:
  Data Mining, Inference, and Prediction}\/} (pp. \bibinfo{pages}{295--336}).
\newblock \bibinfo{address}{New York, NY}: \bibinfo{publisher}{Springer New
  York}.
\bibitem[{Hastie et~al.(2009{\natexlab{b}})Hastie, Tibshirani \&
  Friedman}]{suppor_vector_machines}
\bibinfo{author}{Hastie, T.}, \bibinfo{author}{Tibshirani, R.}, \&
  \bibinfo{author}{Friedman, J.} (\bibinfo{year}{2009}{\natexlab{b}}).
\newblock \bibinfo{title}{Support vector machines and flexible discriminants}.
\newblock In {\it \bibinfo{booktitle}{The Elements of Statistical Learning:
  Data Mining, Inference, and Prediction}\/} (pp. \bibinfo{pages}{417--458}).
\newblock \bibinfo{address}{New York, NY}: \bibinfo{publisher}{Springer New
  York}.
\newblock \DOIprefix\doi{10.1007/978-0-387-84858-7_12}.
\bibitem[{Hauck(2017)}]{Pizzagate}
\bibinfo{author}{Hauck, G.} (\bibinfo{year}{2017}).
\newblock \bibinfo{title}{\textit{\textquotesingle
  \capitalisewords{pizzagate}\textquotesingle\ shooter sentenced to 4 years in
  prison}}.
\newblock \bibinfo{howpublished}{CNN}.
\newblock
  \bibinfo{note}{\url{https://edition.cnn.com/2017/06/22/politics/pizzagate-sentencing/index.html}}.
\bibitem[{Hochreiter \& Schmidhuber(1997)}]{LSTM_network}
\bibinfo{author}{Hochreiter, S.}, \& \bibinfo{author}{Schmidhuber, J.}
  (\bibinfo{year}{1997}).
\newblock \bibinfo{title}{{Long Short-Term Memory}}.
\newblock {\it \bibinfo{journal}{Neural Computation}\/},  {\it
  \bibinfo{volume}{9}\/}, \bibinfo{pages}{1735--1780}.
  \DOIprefix\doi{10.1162/neco.1997.9.8.1735}.
\bibitem[{Hossain et~al.(2019)Hossain, Sohel, Shiratuddin \&
  Laga}]{related_work_Hossain}
\bibinfo{author}{Hossain, M.~Z.}, \bibinfo{author}{Sohel, F.},
  \bibinfo{author}{Shiratuddin, M.~F.}, \& \bibinfo{author}{Laga, H.}
  (\bibinfo{year}{2019}).
\newblock \bibinfo{title}{A comprehensive survey of deep learning for image
  captioning}.
\newblock {\it \bibinfo{journal}{ACM Computing Surveys (CsUR)}\/},  {\it
  \bibinfo{volume}{51}\/}, \bibinfo{pages}{1--36}.
  \DOIprefix\doi{https://doi.org/10.1145/3295748}.
\bibitem[{Hunt \& Gentzkow(2017)}]{Allcot_art}
\bibinfo{author}{Hunt, A.}, \& \bibinfo{author}{Gentzkow, M.}
  (\bibinfo{year}{2017}).
\newblock \bibinfo{title}{Social media and fake news in the 2016 election}.
\newblock {\it \bibinfo{journal}{Journal of Economic Perspectives}\/},  {\it
  \bibinfo{volume}{31}\/}, \bibinfo{pages}{211--236}.
  \DOIprefix\doi{10.1257/jep.31.2.211}.
\bibitem[{Islam et~al.(2021)Islam, Kamal, Kabir, Southern, Khan, Hasan, Sarkar,
  Sharmin, Das, Roy, Harun, Chughtai, Homaira \& Seale}]{Vaccine_hesitancy}
\bibinfo{author}{Islam, M.}, \bibinfo{author}{Kamal, A.-H.},
  \bibinfo{author}{Kabir, A.}, \bibinfo{author}{Southern, D.},
  \bibinfo{author}{Khan, S.}, \bibinfo{author}{Hasan, S.},
  \bibinfo{author}{Sarkar, T.}, \bibinfo{author}{Sharmin, S.},
  \bibinfo{author}{Das, S.}, \bibinfo{author}{Roy, T.}, \bibinfo{author}{Harun,
  G.~D.}, \bibinfo{author}{Chughtai, A.}, \bibinfo{author}{Homaira, N.}, \&
  \bibinfo{author}{Seale, H.} (\bibinfo{year}{2021}).
\newblock \bibinfo{title}{\uppercase{COVID}-19 vaccine rumors and conspiracy
  theories: The need for cognitive inoculation against misinformation to
  improve vaccine adherence}.
\newblock {\it \bibinfo{journal}{PLOS ONE}\/},  {\it \bibinfo{volume}{16}\/},
  \bibinfo{pages}{Article e0251605}.
  \DOIprefix\doi{10.1371/journal.pone.0251605}.
\bibitem[{Jin et~al.(2017)Jin, Cao, Guo, Zhang \& Luo}]{related_work_Jin}
\bibinfo{author}{Jin, Z.}, \bibinfo{author}{Cao, J.}, \bibinfo{author}{Guo,
  H.}, \bibinfo{author}{Zhang, Y.}, \& \bibinfo{author}{Luo, J.}
  (\bibinfo{year}{2017}).
\newblock \bibinfo{title}{Multimodal fusion with recurrent neural networks for
  rumor detection on microblogs}.
\newblock In {\it \bibinfo{booktitle}{{Proceedings of the 25th ACM
  international conference on Multimedia}}\/} (pp. \bibinfo{pages}{795--816}).
\bibitem[{Kaggle({\natexlab{a}})}]{related_work_Kaggle_BuzzFeed}
\bibinfo{author}{Kaggle} ({\natexlab{a}}).
\newblock \bibinfo{title}{{FakeNewsNet}}.
\newblock \bibinfo{note}{Retrieved from
  \url{https://www.kaggle.com/mdepak/fakenewsnet}. Accessed October 14, 2021}.
\bibitem[{Kaggle({\natexlab{b}})}]{related_work_Kaggle_dataset}
\bibinfo{author}{Kaggle} ({\natexlab{b}}).
\newblock \bibinfo{title}{{Getting Real about Fake News}}.
\newblock \bibinfo{note}{Retrieved from
  \url{https://www.kaggle.com/mrisdal/fake-news}. Accessed October 13, 2021}.
\bibitem[{Kaliyar et~al.(2020)Kaliyar, Kumar, Kumar, Narkhede, Namboodiri \&
  Mishra}]{related_work_Kaliyar}
\bibinfo{author}{Kaliyar, R.~K.}, \bibinfo{author}{Kumar, P.},
  \bibinfo{author}{Kumar, M.}, \bibinfo{author}{Narkhede, M.},
  \bibinfo{author}{Namboodiri, S.}, \& \bibinfo{author}{Mishra, S.}
  (\bibinfo{year}{2020}).
\newblock \bibinfo{title}{{DeepNet: An Efficient Neural Network for Fake News
  Detection using News-User Engagements}}.
\newblock In {\it \bibinfo{booktitle}{2020 5th International Conference on
  Computing, Communication and Security (ICCCS)}\/} (pp.
  \bibinfo{pages}{1--6}).
\newblock \DOIprefix\doi{10.1109/ICCCS49678.2020.9277353}.
\bibitem[{Kang et~al.(2021)Kang, Cao, Shang, Liang, Tang \&
  Tong}]{related_work_Kang}
\bibinfo{author}{Kang, Z.}, \bibinfo{author}{Cao, Y.}, \bibinfo{author}{Shang,
  Y.}, \bibinfo{author}{Liang, T.}, \bibinfo{author}{Tang, H.}, \&
  \bibinfo{author}{Tong, L.} (\bibinfo{year}{2021}).
\newblock \bibinfo{title}{Fake news detection with heterogenous deep graph
  convolutional network}.
\newblock In \bibinfo{editor}{K.~Karlapalem}, \bibinfo{editor}{H.~Cheng},
  \bibinfo{editor}{N.~Ramakrishnan}, \bibinfo{editor}{R.~K. Agrawal},
  \bibinfo{editor}{P.~K. Reddy}, \bibinfo{editor}{J.~Srivastava}, \&
  \bibinfo{editor}{T.~Chakraborty} (Eds.), {\it \bibinfo{booktitle}{Advances in
  Knowledge Discovery and Data Mining}\/} (pp. \bibinfo{pages}{408--420}).
\newblock \bibinfo{address}{Cham}: \bibinfo{publisher}{Springer International
  Publishing}.
\bibitem[{Kingma \& Ba(2014)}]{methods_Kingma}
\bibinfo{author}{Kingma, D.~P.}, \& \bibinfo{author}{Ba, J.}
  (\bibinfo{year}{2014}).
\newblock \bibinfo{title}{Adam: A method for stochastic optimization}.
\newblock {\it \bibinfo{journal}{arXiv preprint arXiv:1412.6980}\/}, .
\bibitem[{Kirchknopf et~al.(2021)Kirchknopf, Slijepcevic \&
  Zeppelzauer}]{related_work_Kirchknopf}
\bibinfo{author}{Kirchknopf, A.}, \bibinfo{author}{Slijepcevic, D.}, \&
  \bibinfo{author}{Zeppelzauer, M.} (\bibinfo{year}{2021}).
\newblock \bibinfo{title}{{Multimodal Detection of Information Disorder from
  Social Media}}.
\newblock {\it \bibinfo{journal}{arXiv preprint arXiv:2105.15165}\/}, .
\bibitem[{Kleinbaum \& Klein(2010)}]{logistic_regression}
\bibinfo{author}{Kleinbaum, D.~G.}, \& \bibinfo{author}{Klein, M.}
  (\bibinfo{year}{2010}).
\newblock {\it \bibinfo{title}{Logistic Regression}\/}.
\newblock (\bibinfo{edition}{3rd} ed.).
\newblock \bibinfo{publisher}{\capitalisewords{Springer-Verlag New York}}.
\bibitem[{Kumari \& Ekbal(2021)}]{kumari_fake_news}
\bibinfo{author}{Kumari, R.}, \& \bibinfo{author}{Ekbal, A.}
  (\bibinfo{year}{2021}).
\newblock \bibinfo{title}{{AMFB: Attention based multimodal Factorized Bilinear
  Pooling for multimodal Fake News Detection}}.
\newblock {\it \bibinfo{journal}{Expert Systems with Applications}\/},  {\it
  \bibinfo{volume}{184}\/}, \bibinfo{pages}{115412}.
  \DOIprefix\doi{https://doi.org/10.1016/j.eswa.2021.115412}.
\bibitem[{Li et~al.(2021)Li, Sun, Yu, Tian, Yao \& Xu}]{related_work_Li2}
\bibinfo{author}{Li, P.}, \bibinfo{author}{Sun, X.}, \bibinfo{author}{Yu, H.},
  \bibinfo{author}{Tian, Y.}, \bibinfo{author}{Yao, F.}, \&
  \bibinfo{author}{Xu, G.} (\bibinfo{year}{2021}).
\newblock \bibinfo{title}{{Entity-Oriented Multi-Modal Alignment and Fusion
  Network for Fake News Detection}}.
\newblock {\it \bibinfo{journal}{IEEE Transactions on Multimedia}\/},  (pp.
  \bibinfo{pages}{1--1}). \DOIprefix\doi{10.1109/TMM.2021.3098988}.
\bibitem[{Minaee et~al.(2021{\natexlab{a}})Minaee, Kalchbrenner, Cambria,
  Nikzad, Chenaghlu \& Gao}]{minaee2021deep}
\bibinfo{author}{Minaee, S.}, \bibinfo{author}{Kalchbrenner, N.},
  \bibinfo{author}{Cambria, E.}, \bibinfo{author}{Nikzad, N.},
  \bibinfo{author}{Chenaghlu, M.}, \& \bibinfo{author}{Gao, J.}
  (\bibinfo{year}{2021}{\natexlab{a}}).
\newblock \bibinfo{title}{Deep learning--based text classification: A
  comprehensive review}.
\newblock {\it \bibinfo{journal}{ACM Computing Surveys (CSUR)}\/},  {\it
  \bibinfo{volume}{54}\/}, \bibinfo{pages}{1--40}.
  \DOIprefix\doi{https://doi.org/10.1145/3439726}.
\bibitem[{Minaee et~al.(2021{\natexlab{b}})Minaee, Kalchbrenner, Cambria,
  Nikzad, Chenaghlu \& Gao}]{methods_Minaee}
\bibinfo{author}{Minaee, S.}, \bibinfo{author}{Kalchbrenner, N.},
  \bibinfo{author}{Cambria, E.}, \bibinfo{author}{Nikzad, N.},
  \bibinfo{author}{Chenaghlu, M.}, \& \bibinfo{author}{Gao, J.}
  (\bibinfo{year}{2021}{\natexlab{b}}).
\newblock \bibinfo{title}{{Deep Learning--based Text Classification: A
  Comprehensive Review}}.
\newblock {\it \bibinfo{journal}{ACM Computing Surveys (CSUR)}\/},  {\it
  \bibinfo{volume}{54}\/}, \bibinfo{pages}{1--40}.
  \DOIprefix\doi{https://doi.org/10.1145/3439726}.
\bibitem[{Mishra(2019)}]{India_lynching}
\bibinfo{author}{Mishra, V.} (\bibinfo{year}{2019}).
\newblock \bibinfo{title}{\textit{India’s fake news problem is killing real
  people}}.
\newblock \bibinfo{howpublished}{Asia Times}.
\newblock
  \bibinfo{note}{\url{https://asiatimes.com/2019/10/indias-fake-news-problem-is-killing-real-people/}}.
\bibitem[{Murphy(2012)}]{related_work_Murphy}
\bibinfo{author}{Murphy, K.} (\bibinfo{year}{2012}).
\newblock \bibinfo{title}{Kernels}.
\newblock In {\it \bibinfo{booktitle}{{Machine Learning: A Probabilistic
  Perspective}}\/} (pp. \bibinfo{pages}{479--512}).
\newblock \bibinfo{address}{Cambridge, MA}: \bibinfo{publisher}{The MIT Press}.
\bibitem[{Nakamura et~al.(2020)Nakamura, Levy \& Wang}]{fakkedit}
\bibinfo{author}{Nakamura, K.}, \bibinfo{author}{Levy, S.}, \&
  \bibinfo{author}{Wang, W.~Y.} (\bibinfo{year}{2020}).
\newblock \bibinfo{title}{{F}akeddit: A new multimodal benchmark dataset for
  fine-grained fake news detection}.
\newblock In {\it \bibinfo{booktitle}{Proceedings of the 12th Language
  Resources and Evaluation Conference}\/} (pp. \bibinfo{pages}{6149--6157}).
\newblock \bibinfo{address}{Marseille, France}: \bibinfo{publisher}{European
  Language Resources Association}.
\newblock \URLprefix \url{https://aclanthology.org/2020.lrec-1.755}.
\bibitem[{O’Mahony et~al.(2019)O’Mahony, Campbell, Carvalho, Harapanahalli,
  Hernandez, Krpalkova, Riordan \& Walsh}]{related_work_Mahony}
\bibinfo{author}{O’Mahony, N.}, \bibinfo{author}{Campbell, S.},
  \bibinfo{author}{Carvalho, A.}, \bibinfo{author}{Harapanahalli, S.},
  \bibinfo{author}{Hernandez, G.~V.}, \bibinfo{author}{Krpalkova, L.},
  \bibinfo{author}{Riordan, D.}, \& \bibinfo{author}{Walsh, J.}
  (\bibinfo{year}{2019}).
\newblock \bibinfo{title}{Deep learning vs. traditional computer vision}.
\newblock In {\it \bibinfo{booktitle}{Science and Information Conference}\/}
  (pp. \bibinfo{pages}{128--144}).
\newblock \bibinfo{organization}{Springer}.
\bibitem[{Patel \& Bhattacharyya(2017)}]{methods_Patel}
\bibinfo{author}{Patel, K.}, \& \bibinfo{author}{Bhattacharyya, P.}
  (\bibinfo{year}{2017}).
\newblock \bibinfo{title}{Towards lower bounds on number of dimensions for word
  embeddings}.
\newblock In {\it \bibinfo{booktitle}{Proceedings of the Eighth International
  Joint Conference on Natural Language Processing (Volume 2: Short Papers)}\/}
  (pp. \bibinfo{pages}{31--36}).
\bibitem[{Patwa et~al.(2020)Patwa, Sharma, Pykl, Guptha, Kumari, Akhtar, Ekbal,
  Das \& Chakraborty}]{related_work_Patwa}
\bibinfo{author}{Patwa, P.}, \bibinfo{author}{Sharma, S.},
  \bibinfo{author}{Pykl, S.}, \bibinfo{author}{Guptha, V.},
  \bibinfo{author}{Kumari, G.}, \bibinfo{author}{Akhtar, M.~S.},
  \bibinfo{author}{Ekbal, A.}, \bibinfo{author}{Das, A.}, \&
  \bibinfo{author}{Chakraborty, T.} (\bibinfo{year}{2020}).
\newblock \bibinfo{title}{{Fighting an infodemic: Covid-19 fake news dataset}}.
\newblock {\it \bibinfo{journal}{arXiv preprint arXiv:2011.03327}\/}, .
\bibitem[{Pennington et~al.(2014)Pennington, Socher \& Manning}]{methods_GloVe}
\bibinfo{author}{Pennington, J.}, \bibinfo{author}{Socher, R.}, \&
  \bibinfo{author}{Manning, C.~D.} (\bibinfo{year}{2014}).
\newblock \bibinfo{title}{Glove: Global vectors for word representation}.
\newblock In {\it \bibinfo{booktitle}{Proceedings of the 2014 conference on
  empirical methods in natural language processing (EMNLP)}\/} (pp.
  \bibinfo{pages}{1532--1543}).
\bibitem[{Sabour et~al.(2017)Sabour, Frosst \& Hinton}]{related_work_Sabour}
\bibinfo{author}{Sabour, S.}, \bibinfo{author}{Frosst, N.}, \&
  \bibinfo{author}{Hinton, G.~E.} (\bibinfo{year}{2017}).
\newblock \bibinfo{title}{Dynamic routing between capsules}.
\newblock {\it \bibinfo{journal}{arXiv preprint arXiv:1710.09829}\/}, .
\bibitem[{Sanh et~al.(2019)Sanh, Debut, Chaumond \& Wolf}]{related_work_Sanh}
\bibinfo{author}{Sanh, V.}, \bibinfo{author}{Debut, L.},
  \bibinfo{author}{Chaumond, J.}, \& \bibinfo{author}{Wolf, T.}
  (\bibinfo{year}{2019}).
\newblock \bibinfo{title}{Distilbert, a distilled version of bert: smaller,
  faster, cheaper and lighter}.
\newblock {\it \bibinfo{journal}{arXiv preprint arXiv:1910.01108}\/}, .
\bibitem[{Sharma \& Gupta(2018)}]{related_work_Sharma}
\bibinfo{author}{Sharma, Y.}, \& \bibinfo{author}{Gupta, S.}
  (\bibinfo{year}{2018}).
\newblock \bibinfo{title}{Deep learning approaches for question answering
  system}.
\newblock {\it \bibinfo{journal}{Procedia computer science}\/},  {\it
  \bibinfo{volume}{132}\/}, \bibinfo{pages}{785--794}.
  \DOIprefix\doi{https://doi.org/10.1016/j.procs.2018.05.090}.
\bibitem[{Shu et~al.(2020)Shu, Mahudeswaran, Wang, Lee \&
  Liu}]{related_work_Shu}
\bibinfo{author}{Shu, K.}, \bibinfo{author}{Mahudeswaran, D.},
  \bibinfo{author}{Wang, S.}, \bibinfo{author}{Lee, D.}, \&
  \bibinfo{author}{Liu, H.} (\bibinfo{year}{2020}).
\newblock \bibinfo{title}{{FakeNewsNet: A Data Repository with News Content,
  Social Context, and Spatiotemporal Information for Studying Fake News on
  Social Media}}.
\newblock {\it \bibinfo{journal}{Big data}\/},  {\it \bibinfo{volume}{8}\/},
  \bibinfo{pages}{171--188}. \DOIprefix\doi{10.1089/big.2020.0062}.
\bibitem[{Simonyan \& Zisserman(2014)}]{related_work_Simonyan}
\bibinfo{author}{Simonyan, K.}, \& \bibinfo{author}{Zisserman, A.}
  (\bibinfo{year}{2014}).
\newblock \bibinfo{title}{Very deep convolutional networks for large-scale
  image recognition}.
\newblock {\it \bibinfo{journal}{arXiv preprint arXiv:1409.1556}\/}, .
\bibitem[{Singh et~al.(2021)Singh, Ghosh \& Sonagara}]{singh_fake_news}
\bibinfo{author}{Singh, V.~K.}, \bibinfo{author}{Ghosh, I.}, \&
  \bibinfo{author}{Sonagara, D.} (\bibinfo{year}{2021}).
\newblock \bibinfo{title}{Detecting fake news stories via multimodal analysis}.
\newblock {\it \bibinfo{journal}{Journal of the Association for Information
  Science and Technology}\/},  {\it \bibinfo{volume}{72}\/},
  \bibinfo{pages}{3--17}.
\bibitem[{Singhal et~al.(2019)Singhal, Shah, Chakraborty, Kumaraguru \&
  Satoh}]{singhal_fake_news}
\bibinfo{author}{Singhal, S.}, \bibinfo{author}{Shah, R.~R.},
  \bibinfo{author}{Chakraborty, T.}, \bibinfo{author}{Kumaraguru, P.}, \&
  \bibinfo{author}{Satoh, S.} (\bibinfo{year}{2019}).
\newblock \bibinfo{title}{Spot\uppercase{F}ake: A multi-modal framework for
  fake news detection}.
\newblock In {\it \bibinfo{booktitle}{2019 IEEE fifth international conference
  on multimedia big data (BigMM)}\/} (pp. \bibinfo{pages}{39--47}).
\newblock \bibinfo{organization}{IEEE}.
\bibitem[{Tang et~al.(2015)Tang, Qin \& Liu}]{related_work_Tang}
\bibinfo{author}{Tang, D.}, \bibinfo{author}{Qin, B.}, \& \bibinfo{author}{Liu,
  T.} (\bibinfo{year}{2015}).
\newblock \bibinfo{title}{Deep learning for sentiment analysis: successful
  approaches and future challenges}.
\newblock {\it \bibinfo{journal}{Wiley Interdisciplinary Reviews: Data Mining
  and Knowledge Discovery}\/},  {\it \bibinfo{volume}{5}\/},
  \bibinfo{pages}{292--303}. \DOIprefix\doi{https://doi.org/10.1002/widm.1171}.
\bibitem[{Thota et~al.(2018)Thota, Tilak, Ahluwalia \& Lohia}]{thota_fake_news}
\bibinfo{author}{Thota, A.}, \bibinfo{author}{Tilak, P.},
  \bibinfo{author}{Ahluwalia, S.}, \& \bibinfo{author}{Lohia, N.}
  (\bibinfo{year}{2018}).
\newblock \bibinfo{title}{Fake news detection: a deep learning approach}.
\newblock {\it \bibinfo{journal}{SMU Data Science Review}\/},  {\it
  \bibinfo{volume}{1}\/}, \bibinfo{pages}{10}.
\bibitem[{Viana et~al.(2017)Viana, Nguyen, Smith \&
  Gabrani}]{related_work_Viana}
\bibinfo{author}{Viana, M.}, \bibinfo{author}{Nguyen, Q.-B.},
  \bibinfo{author}{Smith, J.}, \& \bibinfo{author}{Gabrani, M.}
  (\bibinfo{year}{2017}).
\newblock \bibinfo{title}{{Multimodal Classification of Document Embedded
  Images}}.
\newblock In {\it \bibinfo{booktitle}{International Workshop on Graphics
  Recognition}\/} (pp. \bibinfo{pages}{45--53}).
\newblock \bibinfo{organization}{Springer}.
\bibitem[{Voita(2021)}]{methods_Voita}
\bibinfo{author}{Voita, E.} (\bibinfo{year}{2021}).
\newblock \bibinfo{title}{{Convolutional Neural Networks for Text}}.
\newblock
  \bibinfo{howpublished}{\url{https://lena-voita.github.io/nlp_course/models/convolutional.html}}.
\newblock \bibinfo{note}{Accessed October 5, 2021}.
\bibitem[{Wang(2017)}]{related_work_Wang_LIAR}
\bibinfo{author}{Wang, W.~Y.} (\bibinfo{year}{2017}).
\newblock \bibinfo{title}{{"Liar, Liar Pants on Fire": A new benchmark dataset
  for fake news detection}}.
\newblock {\it \bibinfo{journal}{arXiv preprint arXiv:1705.00648}\/}, .
\bibitem[{Wang et~al.(2018)Wang, Ma, Jin, Yuan, Xun, Jha, Su \&
  Gao}]{wang_fake_news}
\bibinfo{author}{Wang, Y.}, \bibinfo{author}{Ma, F.}, \bibinfo{author}{Jin,
  Z.}, \bibinfo{author}{Yuan, Y.}, \bibinfo{author}{Xun, G.},
  \bibinfo{author}{Jha, K.}, \bibinfo{author}{Su, L.}, \& \bibinfo{author}{Gao,
  J.} (\bibinfo{year}{2018}).
\newblock \bibinfo{title}{\uppercase{EANN}: Event adversarial neural networks
  for multi-modal fake news detection}.
\newblock In {\it \bibinfo{booktitle}{Proceedings of the 24th \uppercase{acm
  sigkdd} international conference on knowledge discovery \& data mining}\/}
  (pp. \bibinfo{pages}{849--857}).
\bibitem[{Wani et~al.(2021)Wani, Joshi, Khandve, Wagh \&
  Joshi}]{related_work_Wani}
\bibinfo{author}{Wani, A.}, \bibinfo{author}{Joshi, I.},
  \bibinfo{author}{Khandve, S.}, \bibinfo{author}{Wagh, V.}, \&
  \bibinfo{author}{Joshi, R.} (\bibinfo{year}{2021}).
\newblock \bibinfo{title}{{Evaluating deep learning approaches for Covid19 fake
  news detection}}.
\newblock In {\it \bibinfo{booktitle}{Combating Online Hostile Posts in
  Regional Languages during Emergency Situation: First International Workshop,
  CONSTRAINT 2021, Collocated with AAAI 2021, Virtual Event, February 8, 2021,
  Revised Selected Papers}\/} (p. \bibinfo{pages}{153}).
\newblock \bibinfo{organization}{Springer Nature}.
\bibitem[{Xie et~al.(2021)Xie, Liu, Liu, Zhang \& Zhu}]{related_work_Xie}
\bibinfo{author}{Xie, J.}, \bibinfo{author}{Liu, S.}, \bibinfo{author}{Liu,
  R.}, \bibinfo{author}{Zhang, Y.}, \& \bibinfo{author}{Zhu, Y.}
  (\bibinfo{year}{2021}).
\newblock \bibinfo{title}{{SERN: Stance Extraction and Reasoning Network for
  Fake News Detection}}.
\newblock In {\it \bibinfo{booktitle}{{ICASSP 2021 - 2021 IEEE International
  Conference on Acoustics, Speech and Signal Processing (ICASSP)}}\/} (pp.
  \bibinfo{pages}{2520--2524}).
\newblock \DOIprefix\doi{10.1109/ICASSP39728.2021.9414787}.
\bibitem[{Yang et~al.(2016)Yang, Yang, Dyer, He, Smola \&
  Hovy}]{related_work_Yang}
\bibinfo{author}{Yang, Z.}, \bibinfo{author}{Yang, D.}, \bibinfo{author}{Dyer,
  C.}, \bibinfo{author}{He, X.}, \bibinfo{author}{Smola, A.}, \&
  \bibinfo{author}{Hovy, E.} (\bibinfo{year}{2016}).
\newblock \bibinfo{title}{Hierarchical attention networks for document
  classification}.
\newblock In {\it \bibinfo{booktitle}{{Proceedings of the 2016 conference of
  the North American chapter of the association for computational linguistics:
  human language technologies}}\/} (pp. \bibinfo{pages}{1480--1489}).
\bibitem[{Yu \& Jiang(2019)}]{relatead_work_Yu}
\bibinfo{author}{Yu, J.}, \& \bibinfo{author}{Jiang, J.}
  (\bibinfo{year}{2019}).
\newblock \bibinfo{title}{{Adapting BERT for Target-Oriented Multimodal
  Sentiment Classification}}.
\newblock In {\it \bibinfo{booktitle}{Proceedings of the Twenty-Eighth
  International Joint Conference on Artificial Intelligence, {IJCAI-19}}\/}
  (pp. \bibinfo{pages}{5408--5414}).
\bibitem[{Zhao et~al.(2019)Zhao, Zheng, Xu \& Wu}]{related_work_Zhao}
\bibinfo{author}{Zhao, Z.-Q.}, \bibinfo{author}{Zheng, P.},
  \bibinfo{author}{Xu, S.-T.}, \& \bibinfo{author}{Wu, X.}
  (\bibinfo{year}{2019}).
\newblock \bibinfo{title}{Object detection with deep learning: A review}.
\newblock {\it \bibinfo{journal}{IEEE transactions on neural networks and
  learning systems}\/},  {\it \bibinfo{volume}{30}\/},
  \bibinfo{pages}{3212--3232}. \DOIprefix\doi{10.1109/TNNLS.2018.2876865}.
\bibitem[{Zubiaga et~al.(2017)Zubiaga, Liakata \&
  Procter}]{related_work_Zubiaga_PHEME}
\bibinfo{author}{Zubiaga, A.}, \bibinfo{author}{Liakata, M.}, \&
  \bibinfo{author}{Procter, R.} (\bibinfo{year}{2017}).
\newblock \bibinfo{title}{Exploiting context for rumour detection in social
  media}.
\newblock In {\it \bibinfo{booktitle}{International Conference on Social
  Informatics}\/} (pp. \bibinfo{pages}{109--123}).
\newblock \bibinfo{organization}{Springer}.

\end{thebibliography}

\end{document}